\documentclass[conference]{IEEEtran}

\usepackage{balance}

\IEEEoverridecommandlockouts 
\usepackage[dvips]{graphicx}
\usepackage{algorithmic}
\usepackage{algorithm}
\usepackage{listings}
\usepackage[OT4,T1]{fontenc}
\usepackage[cmex10]{amsmath}
\usepackage{url}
\usepackage{multirow}
\usepackage[hidelinks]{hyperref}

\usepackage{cite}
\usepackage{float}
\usepackage{tikz}
\usepackage{caption}
\usepackage{subcaption}
\usetikzlibrary{shadows}
\usetikzlibrary{arrows}
\usetikzlibrary{patterns}
\usepackage{physics}
\usepackage[cp1250]
{inputenc}

\newlength\myindent
\newtheorem{remark}{Remark}[section]

\newtheorem{theorem}{Theorem}

\title{Enhancing naive classifier for positive unlabeled data  based on logistic regression approach}
\author{
\IEEEauthorblockN{Mateusz P{\l}atek}
\IEEEauthorblockA{
Warsaw University of Technology\\
Faculty of Mathematics and Information Science\\
Koszykowa 75, 00-662 Warsaw, Poland\\
Email: mateusz.platek.student@pw.edu.pl}
\and
\IEEEauthorblockN{Jan Mielniczuk}
\IEEEauthorblockA{Institute of Computer Science\\
Polish Academy of Sciences\\
Jana Kazimierza 5, 01-248 Warsaw, Poland\\
and\\
Warsaw University of Technology\\
Faculty of Mathematics and Information Science\\
Koszykowa 75, 00-662 Warsaw, Poland\\
Email: jan.mielniczuk@ipipan.waw.pl}
}

\begin{document}
\maketitle              

\begin{abstract}
We argue that for analysis of Positive Unlabeled (PU) data under Selected Completely At Random (SCAR) assumption it is fruitful to view the problem as  fitting of misspecified model to the data. Namely, we show that the results on misspecified fit imply that in the case when posterior probability of the response is modelled by logistic regression, fitting  the logistic regression to the observable PU data  which {\it does not} follow this model, still yields the vector of estimated parameters approximately colinear with  the true vector of parameters.  This observation  together with choosing the intercept of the classifier based on optimisation of analogue of F1 measure yields a classifier which performs on par or better than its competitors on several real data sets considered.
\end{abstract}

\section{Introduction}
\IEEEoverridecommandlockouts\IEEEPARstart{I}n the paper we analyse classification problem for partially observable data scenario  for which in the case of  some observations   class indicators assigned to them (positive or negative in the case of binary classification) are unknown. More specifically, for positive and unlabeled data considered here, it is assumed that some observations from the positive class are labeled, whereas the rest of of the observations (either positive or negative) are unlabeled. Such scenario is called Positive Unlabelled (PU) scenario.
Thus in the PU setting the true binary class indicator $Y\in \{0,1\}$ is not observed directly but only through binary label  $S$. One knows that if $S=1$ (labelled case),
$Y$ has to be 1 (positive), but for $S=0$ (unlabeled case) $Y$ may be either 1 or 0 (positive or negative). Besides, each object is described by the vector of features $x$. This setup encompasses a  legion of practical situations, in which effective inference methods  about class indicator $Y$ are sought. Examples include disease data (diagnosed patients with a specific   disease detected, and patients yet to be diagnosed who may be ill or not), web pages preferences of a  specific  user (pages bookmarked as of interest   and 
 pages not yet viewed, thus of unknown interest) and ecological examples when environments are  labeled  provided   a specific specimen inhabits  them, and unlabeled,  where this specimen has not been yet looked for). Such scenario is also relevant for survey data, when questions concerning socially reproachable behaviour may not be answered truthfully.\\
One of the popular approaches to learn from PU data is to impose certain parametric assumptions on distribution of $(X,Y)$ as it commonly done in classical classification task together with some assumptions on labeling mechanism $S$. This is partly necessitated by the  fact that  in general situation the posterior  distribution of $Y$  as well as prior probability $P(Y=1)$ is not identifiable.  It is thus common to consider logistic type of dependence for the posterior distribution $P(Y=1|X=x)$ and assume that censoring mechanism acts indiscriminately of $x$ and is described only by the label frequency $c=P(S=1|Y=1)$ (SCAR assumption discussed below). Majority of learning approaches has been developed 
 under such assumptions; see \cite{BekkerSurvey} for an extensive review of the proposed methods.
Recently the JOINT method has been proposed in \cite{PUICCS}  which consists in minimisation of empirical risk for the observed data $(X_i,S_i),i=1,\ldots,n$ with respect to parameter of logistic distribution {\it and} label frequency. JOINT method can be considered as a generic method with specific algorithms depending on optimisation technique used. The issue is delicate as it turns out that the empirical risk is {\it not} a convex function of its parameters and thus it may posess multiple local minima. In particular \cite{PUICCS} used BFGS algorithm, whereas approach in  \cite{MM2021} has been based on Minorization-Maximization (MM) technique. Among other methods important group consists of approaches based on weighted empirical risk minimisation in which weights of observations depend on  labeling frequency $c$ (see \cite{BekkerSurvey}, section 5.3.2).\\
In the present contribution we  call attention to the fact  that in order to construct a  reasonable classifier one can use a logistic model fitted to observable data $(X_i,S_i), i=1,\ldots,n$ in order to recover the direction of the separating hyperplane and then shift it to the optimal position by maximising observable analogue of $F1$ score. In this approach  the direction is obtained by minimising the misspecified {\it convex}  empirical risk (equal to minus log-likelihood) for the observed data.
The justification of the method is based on properties of misspecified logistic regression  which are valid for PU model under SCAR  condition considered here. We argue that considering  fitting parametric models to PU data  as the  misspecification problem gives new insights  to the established properties and leads to  new solutions. In particular, results on behaviour of estimators under misspecification (see e.g. \cite{White1982}, \cite{Vuong}) can be used to assess the performance of the naive classifier and its modifications.

\section{Notions and auxiliary results}
\label{prelims}
We first introduce basic notations. Let $X$ be a  multivariate random variable corresponding to feature vector, $Y\in\{0,1\}$ be a true class label and $S\in\{0,1\}$ an indicator of  an example being labeled ($S=1$) or not ($S=0$). 
We consider $X$ as a column vector and let $X=(1,\tilde X^T)^T\in R^{p+1}$, where the first coordinate of $X$ corresponds to an intercept and coordinates of $\tilde X$ relate to $p$ collected characteristics of an observation.
We assume that there is some unknown distribution $P_{Y,X,S}$ such that  
$(Y_i,X_i,S_i), i=1,\ldots,n$ is iid sample drawn from it. Observed data  consists of  $(X_i,S_i),i=1,\ldots,n$. This is the single sample scenario as opposed to case-control scenario when the samples from positive class and the general population are given.  
Only positive examples ($Y=1$)  can be labeled, i.e. $P(S=1|X,Y=0)=0$.  Thus we know that $Y=1$ when $S=1$ but when $S=0$,  $Y$ can be either 1 or 0.
Our primary  aim is    is to construct classifier which predicts $Y$ class based on PU data.
Note that this corresponds to a specific censored data problem as  we only   observe samples from distribution of $(X,S)$, where $S=Y$ with a certain probability.

To this end we define  binary posterior probability  of  $S=1$ given $X=x$ equal $s(x)=P(S=1|x)$ and propensity score function $e(x)=P(S=1|Y=1,X=x)$. In this paper we adopt Selected Completely At Random (SCAR) assumption which stipulates that $e(x)$ does not depend on $x$, thus $e(x)=P(S=1|Y=1):=c$, where $c$ will stand for labeling frequency.
This means that labeling is not influenced by  feature vector $x$ and in this case labeled data is a random sample (of a random size) from a  positive class. This commonly adopted assumption is restrictive but it serves as an useful  approximation especially in situations  when the possibility of labeling bias is recognised and  one tries to avoid it.
 We note that as we have $P(S=1,Y=0|X=x)=0$  it holds
\begin{eqnarray}
\label{posteriorS}
s(x)&=&P(S=1|x) =P(S=1,Y=1|x)\cr &=& P(S=1|Y=1,x)P(Y=1|x)\cr &=&
e(x)\times y(x)=c\times y(x),
\end{eqnarray}
where we let $y(x)=P(Y=1|X=x)$  denote posterior probability of class 1 and  the last equality follows from  SCAR assumption.
We note  that  SCAR  is equivalent to the property  that $S$ and $X$ are conditionally independent given $Y$. We stress, however,  that it is valid only when the label value  is assigned with a fixed probability  regardless of characteristics of an item. Under this assumption it is easy to see that $P_{X|S=1}=P_{X|Y=1}$ whereas $P_{X|S=0}$ is a mixture
\[ P_{X|S=0} =\frac{\alpha -\alpha c}{1-\alpha c}P_{X|Y=1} + \frac{1- \alpha }{1-\alpha c}P_{X|Y=0} \]
and  $\alpha=P(Y=1)$ is a prior probability of $Y=1$.
We also note that $c=P(S=1|Y=1)= P(S=1)/P(Y=1)=P(S=1)/\alpha$. We do not assume any previous knowledge of $c$ (although it is frequently imposed see, e.g. \cite{BekkerSurvey}) and thus we only know that $ 0 < c\leq 1$.
We will adopt an parametric model for posterior probability $y(x)$ assuming that 
 $Y$ is governed by logistic response:
\begin{equation}
\label{logistic}
y(x)= \frac{\exp(x^{T}\beta)}{1 +  \exp(x^T\beta)} = \sigma(x^T\beta), 
\end{equation}
where $\sigma(s)=\exp(s)/(1+\exp(s))$ is a logistic function, $\beta^{T}$ stands for  transposed column vector $\beta$ and $\beta=(\beta_0,\beta_{-0}^T)^T\in R\times R^{p}$ is an unknown but fixed vector value.  Thus in view of (\ref{posteriorS}) and (\ref{logistic}) we have 
\[P(S=1|x)=c\times\sigma(x^T\beta).\]
\section{Misspecified logistic modelling}
Assume that (\ref{logistic}) holds and consider naive approach when the logistic model is fitted to $(X,S)$ data using Maximum Likelihood method i.e.  we maximise a log-likelihood
\begin{equation}
\label{mis_emp_risk}
{\cal L}_n(b)= \sum_{i=1}^n S_i\log(\sigma(X_i^Tb)) + (1-S_i)\log(1- \sigma(X_i^Tb)).
\end{equation}
Maximisation of ${\cal L}_n(\cdot)$ is a concave optimisation problem.
Note that this is equivalent to assuming (erroneously) that all  unlabeled observations belong to the negative class and thus misspecified logistic model is fitted to the data for which posterior probability is governed by (\ref{posteriorS}).
Obviously, one can write down the complete correct  log-likelihood for $(X_i,S_i)_{i=1}^n$:
\begin{equation}
\label{emp_risk}
\tilde{\cal L}_n(b,c)= \sum_{i=1}^n S_i\log(c\sigma(X_i^Tb)) + (1-S_i)\log(1- c\sigma(X_i^Tb))
\end{equation}
and maximise  it wrt to $(b,c)$. Such method, named JOINT, has been proposed and investigated  in \cite{PUICCS}. However, finding global  maximum of  (\ref{emp_risk}) is hindered by the fact that due to the presence of  multiplicative constant $c$ in the form of  posterior probability $P(S=1|x)$ given in (\ref{posteriorS}) log-likelihood $\tilde{\cal L}_n(b,c)$ is no longer concave wrt $b$, in contrast to ${\cal L}_n(b)$. There are some attempts to account for this, either by using Minorization-Maximization algorithm or modelling $\tilde{\cal L}_n(\cdot,c)$ as the difference of two concave functions (\cite{AMCS}).\\
Frequently,  our aim is not to approximate $(\beta,c)$ but to construct a classification rule based on training data $(X_i,S_i)_{i=1}^n$. For review of such methods see e.g.  \cite{BekkerSurvey}. 
In such a case one can ask whether the classifier based on maximiser of  ${\cal L}_n(b)$ can not be modified to yield approximation of Bayes classifier of $Y$. The answer is affirmative and it relies on the crucial observation  that  ${\cal L}_n(b)$ can be viewed as log-likelihood of misspecified logistic regression fitted to data corresponding to posterior probability $q(x^T\beta)=c\times \sigma(x^T\beta)$. This was noticed already in the context of estimation of $\beta$ in \cite{PUICCS} using Ruud's theorem \cite{Ruud} stated below, however its useful consequences  have been never explored for PU classification. Here we try to fill this gap by showing that the naive classifier can be improved by adjusting  its intercept, the step which has significant influence on its performance. Below we state Ruud's theorem \cite{Ruud} for a logistic loss, for the general statement see \cite{LiDuan}.\\
\subsection{Colinearity under misspecification: general case}
Assume that the distribution of random vector $(X,S)$ is such that posterior probability 
$P(S=1|X=x)=q(x^T\beta)$ for some unknown  response function $q:R\to (0,1)$ which is possibly different from logistic function.
Let $\beta^*$ be the maximiser of expected normalised log-likelihood in (\ref{mis_emp_risk}) for such distribution:
\begin{eqnarray}
\label{expected_risk}
 n^{-1}E_{(X,S)}{\cal L}_n(b)&=& E_X\{q(x^T\beta)\log\sigma(x^T b) \cr &+&  (1-q(x^T\beta))\log(1-\sigma(x^T b))\} 
\end{eqnarray}

We note that $\beta^*$ can be interpreted as the minimiser of  the averaged Kullback-Leibler (KL) divergence between binary  distribution $(q(X^T\beta), 1-q(X^T\beta)) $ and  family of logistic models $\{\sigma(X^T b)\}_{b\in R^{p+1}}$ (see \cite{CoverThomas} for the definition and properties of KL divergence) and thus corresponds to the Kullback-Leibler projection of the true distribution on this family. 
It also follows that $\beta^*$ satisfies the following vector equality
\begin{equation}
\label{normal_eq}
EXq(X^T\beta)=EX\sigma(X^T\beta^*).
\end{equation}
The  obvious consequence of (\ref{normal_eq}) is  that when $q(s)\equiv \sigma(s)$ and the projection is unique,  then $\beta^*=\beta$. \\
We say that $X$ satisfies Linear Regressions Condition $(LRC(b)$) for vector $b\in R^{p+1}$   if 
\begin{equation}
\label{LRC}
E(\tilde X|\tilde b^T\tilde X=w)=\gamma w+\gamma_0 
\end{equation}
for some $\gamma=\gamma(\tilde b),\gamma_0=\gamma_0(\tilde b)\in R^{p}$.
We note that $LRC(b)$ condition is satisfied for the multivariate normal distribution for any $b\in R^{p+1}$ and, more generally, by  the class of eliptically contoured distributions.

\begin{theorem} \cite{Ruud}
\label{Ruud}
Assume that  $X$ satisfies $LRC(\beta)$ condition  and moreover
covariance matrix of $\Sigma_{\tilde X}$ of vector $\tilde X$ is strictly positive definite.
Additionally,  $P(Y=1|X=x)=q_0(x^T\beta)$ for some unknown function $q_0$  and for some $\beta\in R^{p+1}$.
Then   minimiser $\beta^*$ of (\ref{expected_risk}) satisfies
\[\beta^*_{-0}=\eta \beta_{-0},\] where $\beta=(\beta_0,\beta_{-0}^T)^T$ and $\beta^*=(\beta_0^*,\beta_{-0}^{*^T})^T$.
Moreover,  $\eta>0$ provided that ${\rm Cov}(Y,X)>0$ and $LRC(\beta^*)$ holds.
\end{theorem}

For the proof of the first part see e.g. \cite{LiDuan}.  The second part follows from normal equations (\ref{normal_eq}) and the fact that vector $\gamma$ in (\ref{LRC}) equals 
$(\beta_{-0}^T\Sigma_{\tilde X}\beta_{-0})^{-1}\Sigma_{\tilde X}\beta_{-0}$.

Theorem above implies that 
under the stated conditions despite the misspecification of the fitted model
we still retain colinearity of  true parameter $\beta$ and the vector  $\beta^*$ of its Kullback-Leibler projection  when the first coordinate in both vectors  corresponding to intercept is omitted. This has an obvious relevance in classification if one recalls
that Bayes classifier when logistic model is valid  equals under conditions of Theorem \ref{Ruud}:
\begin{eqnarray}
\label{sep_plane}
& &\hat Y(X)=I\{(\tilde X^{T}\beta_{-0} +\beta_0>0\}\cr&=&I\{\eta\tilde X^T\beta_{-0} +\eta\beta_0> 0\}=
I\{\tilde X^T\beta_{-0}^{*} +\eta\beta_0 > 0\}.
\end{eqnarray}

Thus the direction of the optimal separating hyperplane $\tilde X^T\beta_{-0} +\beta_0=0$ is given by  projection $\beta_{-0}^{*}$ which is easily estimable  and only the intercept $\eta\beta_0$ needs to be recovered.
Let $\hat\beta^*$ denote  maximiser of (\ref{mis_emp_risk}).
As Maximum Likelihood  estimator $\hat\beta^*$  consistently estimates $\beta^*$ under mild conditions (see \cite{White1982}) one can use $\hat\beta^*_{-0}$ as the vector defining the direction of the separating hyperplane $w^Tx +w_{0}$ and then adjust its intercept appropriately.\\
\subsection{Collinearity under misspecification: PU case}
Consider now Positive Unlabeled data case and assume  that posterior probability of $Y$ given $X$  is given by logistic model defined in (\ref{logistic}).
Then in the view of
(\ref{posteriorS}) when logistic model is fitted to $(S,X)$, the model is misspecified as $P(S=1|X=x)=c\times \sigma(^T\beta)$. However, under conditions of Theorem \ref{Ruud} we have $\beta_{-0}^*=\eta\beta_{-0}$ and moreover (\ref{normal_eq}) yields
\[ c EX\sigma(X^T\beta)= EX\sigma(\beta_0^* +\eta\tilde X^T\beta_{-0}).\]
This shows how  parameter $\eta$ depends on  labeling frequency $c$  and distribution of 
$X^T\beta$. When $X$ is multivariate normal  this can be restated more explicitly. 

\begin{theorem}
\label{Additional}
Assume that $X\sim N(0,\Sigma)$ and conditions of Theorem \ref{Ruud} are satisfied. (i) Then we have for any $j=1,\ldots,p$:
\begin{equation}
\label{restated}
\frac{\eta}{c}=\eta\frac{EX_j\sigma(\beta_0 + \tilde X^T\beta_{-0})}{EX_j\sigma(\beta_0^* +\eta\tilde X^T\beta_{-0})}=\frac{E\sigma'(\beta_0 + \tilde X^T\beta_{-0})}{E\sigma'(\beta_0^* +\eta\tilde X^T\beta_{-0})}
\end{equation}
(ii) If $c\leq 1/2$ then $\beta_0^*<0$ for any $\beta_0$.\\
\end{theorem}
Proof. The first equality in (\ref{restated}) is just a consequence of (\ref{normal_eq}) when $j^{th}$ coordinate is considered. The second equality 
 follows from  Stein's lemma, which states that ${\rm Cov}(h(Z_1),Z_2)=Eh'(Z_1){\rm Cov}(Z_1,Z_2)$ for bivariate normal vector $(Z_1,Z_2)$. It implies that
 \begin{eqnarray}
   & &EX_j\sigma(\beta_0 + \tilde X^T\beta_{-0})={\rm Cov}(X_j,\sigma(\beta_0 + \tilde X^T\beta_{-0}))\cr&=&E \sigma'(\beta_0 + \tilde X^T\beta_{-0}){\rm Cov}(X_j,\beta_0 +\tilde X^T\beta_{-0})  
 \end{eqnarray}
and, analogously
\begin{eqnarray}
   & &EX_j\sigma(\beta^*_0 + \eta\tilde X^T\beta^*_{-0})={\rm Cov}(X_j,\sigma(\beta^*_0 + \tilde X^T\beta^*_{-0}))\cr&=&E \sigma'(\beta_0^* +\eta\tilde X^T\beta_{-0}){\rm Cov}(X_j,\beta^*_0 +\eta\tilde X^T\beta^*_{-0}).  
 \end{eqnarray}
 Applying normal equations again we obtain the second equality.\\
 In order to prove (ii) note that
for any symmetric univariate random variable $Z$ we have
\[ E\sigma (a +Z)<1/2 \iff a<0. \]
Indeed 
\[  E\sigma (a +Z) = 1- E\sigma (-a -Z) =1- E\sigma (-a +Z), \]
where the second equation is due to symmetry of $Z$. This,  and the fact that $\sigma(a+Z)<\sigma(-a+Z)$  is equivalent (due to monotonicity of $\sigma(\cdot)$) to $a<0$ justify the claim. However, note that
normal equations for the first coordinate being 1 imply that
\[ E\sigma(\beta_0^* +\eta\tilde X^T\beta_{-0})= cE\sigma(\beta_0 + \tilde X^T\beta_{-0}) <{c}\leq \frac{1}{2} \]
and thus $\beta_0^*<0$. 
\begin{remark}
Part (ii) explains why the naive classifier applied to $(S,X)$ data will work poorly, especially for small $c$: its intercept is likely to be negative regardless the sign of the intercept $\eta\beta_0$ in (\ref{sep_plane}). Thus it has to be modified to enhance the performance of  naive classifier.
\end{remark}
\begin{remark}
The case when no intercept is included in both the true and the fitted model has been considered in \cite{PUICCS}. It is shown there that in then  $0<\eta\leq c<1$. Thus in this case coefficients of logistic model corresponding to genuine predictors are shrunk towards 0.
\end{remark}

\subsection{Choice of the intercept}
We propose to choose the intercept $\widehat w_0$ of the separating hyperplane $\tilde x^T\hat\beta_{-0}^{*} +\widehat w_{0}=0$, where $\widehat w_0$ is an estimator of $\eta\beta_0$ (see (\ref{sep_plane})), by maximising
the analogue of $F1$ measure on training data. We  let, for a given classifier $\hat Y=\hat Y(X)$ learnt on the training data ${\cal D}^{train}$:
\[ r=P(\hat Y(X)=1|Y=1)\quad\quad p=P(Y=1|\hat Y(X)=1)\]
be population recall and precision of $\hat Y$, respectively. Here, $(X,Y)$ stands for unobservable random variable having distribution $P_{X,Y}$ which is independent of ${\cal D}^{train}$.  
We define  $F1$ measure  as their harmonic mean
\begin{equation}
\label{F1}
F1=\frac{r\times p}{(r+p)/2}. 
\end{equation}
Thus in order to have large $F1$ value, both the precision and recall should be large.
We also note that simple derivation yields $ F1=2\times P(Y=1,\hat Y=1)/(P(Y=1) +P(\hat Y=1))$.
Moreover, note that for PU data under SCAR we have that $P(\hat Y(X)=1|Y=1, {\cal D}^{train})=P(\hat Y(X)=1|S=1,{\cal D}^{train})$ as $\hat Y(X)$ given ${\cal D}^{train}$ depends on $X$ only and 
$P(X|Y=1)=P(X|S=1)$. 

This means that  the recall $r$ can be easily estimated from $(X,S)$ sample. The precision, however is unobservable, and thus we consider the following analogue of $F1$ introduced in \cite{LeeLiu03}, Section 4, (see also \cite{Tabata2020}) defined as
\begin{equation}
\label{F1PU}
F1_{PU}=\frac{r\times p}{P(Y=1)}.
\end{equation}
 $F1_{PU}$ is proportional to squared geometric mean of the precision and  the recall i.e. Fowlkes-Mallows index \cite{Fowlkes}. Note that one obtains
\[\frac{P(Y=1|\hat Y(X)=1)}{P(Y=1)}= \frac{P(\hat Y(X)=1|Y=1)}{P(\hat Y(X)=1)} \]
which in terms of the precision and the recall  means that  $p=r\times P(Y=1)/P(\hat Y(X)=1)$ an thus
\begin{equation}
\label{F1PU2}
F1_{PU}=\frac{r^2}{P(\hat Y(X)=1)}.
\end{equation}
Let $\hat Y_z(x)=I\{\tilde x^T\hat{\beta}^{*}_{-0} + z > 0\}$, where $\hat{\beta}^{*}$ is maximiser of (\ref{mis_emp_risk}) and define $\widehat{F1}_{PU}(z)$  to be a sample analogue of
$F1_{PU}$ for the classifier $\hat Y_z(X)$.
We propose  to  choose $\widehat w_0$
as maximiser of 
\begin{equation}
\label{F1PU_max}
  \widehat w_0 ={\rm argmax}_z \widehat{F1}_{PU}(z)  
\end{equation}

We will call the classifier $\hat Y(x)= I\{\tilde x^T\hat\beta_{-0}^{*} +\widehat w_{0}>0\}$ the enhanced naive classifier. The pseudo-code for enhanced classifier is given in Algorithm \ref{alg:F1max}. We show below when analysing its  behaviour on real data sets that modification of the intercept of the naive classifier  is crucial for its performance. 


\begin{algorithm}[]
   \caption{Enhanced naive classifier}
   \label{alg:F1max}
    \begin{algorithmic}[tbp]
        \STATE \textbf{Input:} Observed data $(x_i, s_i)$, $i = 1,\dots, n$. 
        \STATE
        \textbf{Step 1:} Obtain estimator $\hat{\beta}^{*} = (\hat{\beta}^{*}_{0},\hat{\beta}^{*}_{-0})$ by fitting logistic regression to observed data $(x_i, s_i)$.
            \STATE
            \textbf{Step 2:} Calculate intercept $\hat{w}_0$ as $\textrm{argmax}_z \widehat{F1}_{PU}(z)$.
        \STATE 
        \textbf{Result:} Parameters $(\hat{w}_0, \hat{\beta}^{*}_{-0})$ of the separating hyperplane.
    \end{algorithmic}
\end{algorithm}

\section{Numerical experiments}
In the numerical experiments we have considered the following classifiers:
\begin{itemize}
    \item 
Naive classifier based on fitting logistic regression model to $(X,S)$ data called {Naive} and  the classifier Enhanced proposed here;
\item
Classifiers based on JOINT and MM estimators discussed above;
\item
Weighted classifiers introduced in \cite{BekkerSurvey}, Section 5.3.1  using two alternative estimators of $c$: proposed in \cite{ElkanNoto} (denoted by $e_1$, p.214) and TIcE estimator introduced in \cite{TIcE}. For the discussion of both estimators of $c$ see e.g. \cite{MM2021}.
They will be called EN  and TIcE classifiers, respectively.
\end{itemize}
The implementation of Enhanced estimator is given in github directory\footnote{\url{https://github.com/MateuszPlatek/PU_Enhanced_Naive_Classifier} }. Maximisation of $\widehat{F1}_{PU}(z)$ in (\ref{F1PU_max}) is achieved  by  looking  for maximal value among   the  values of this quantity, noting that numerators of numerator and denominator of the ratio defining it may change by $\pm 1$ when moving along ordered values of intercept for which predictions of considered classifiers change, i.e. values $ z_i = \tilde x_i^T\hat{\beta}^{*}_{-0}$.

\subsection{Synthetic data} 
\begin{figure*}[b]
\centering
\hspace*{-8.9cm}
\begin{subfigure}{.6\textwidth}
    \centering  \includegraphics[scale=0.5]{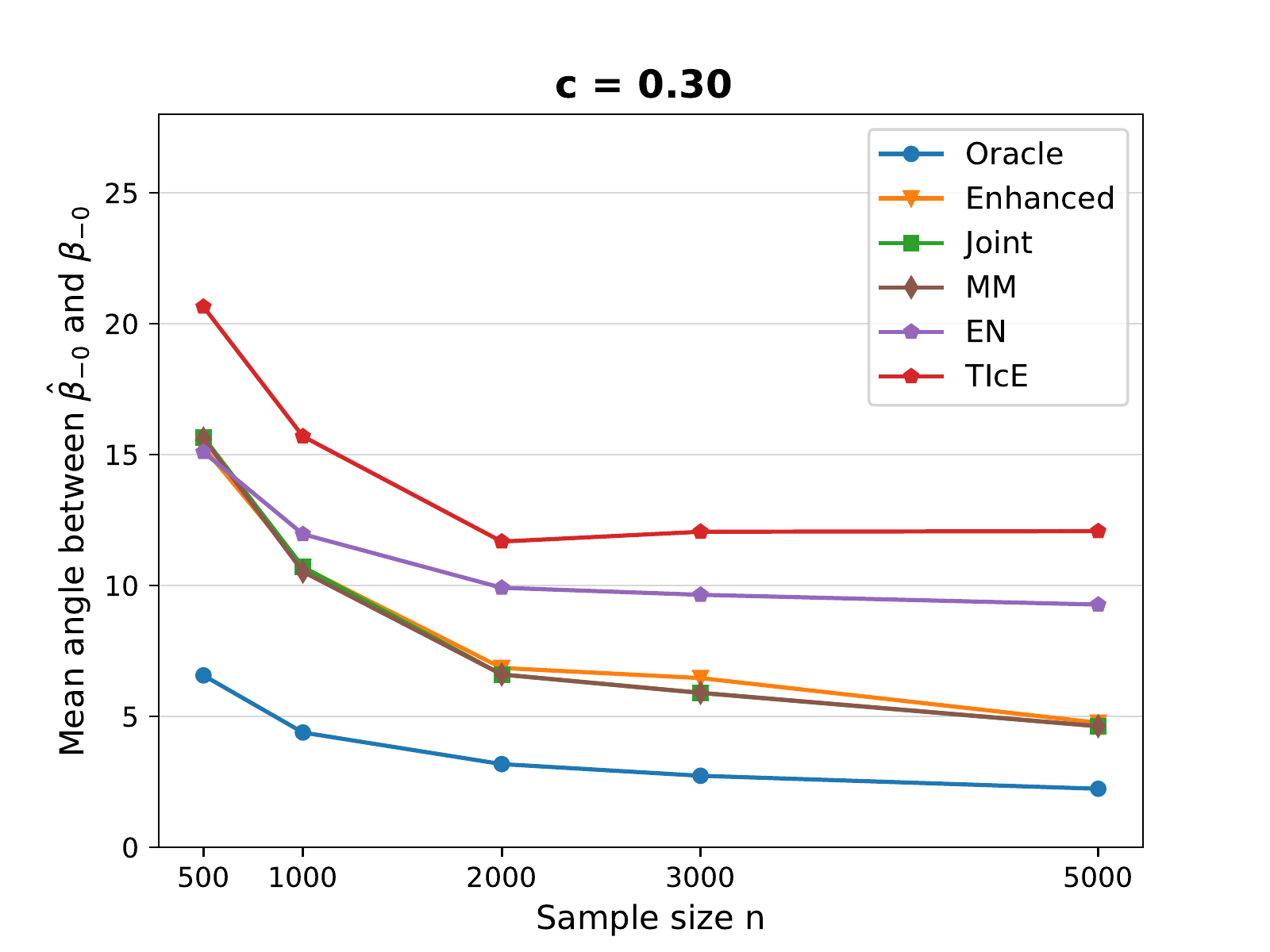}
\end{subfigure}
\hspace*{-2.3cm}
\begin{subfigure}{.01\textwidth}
    \centering \includegraphics[scale=0.5]{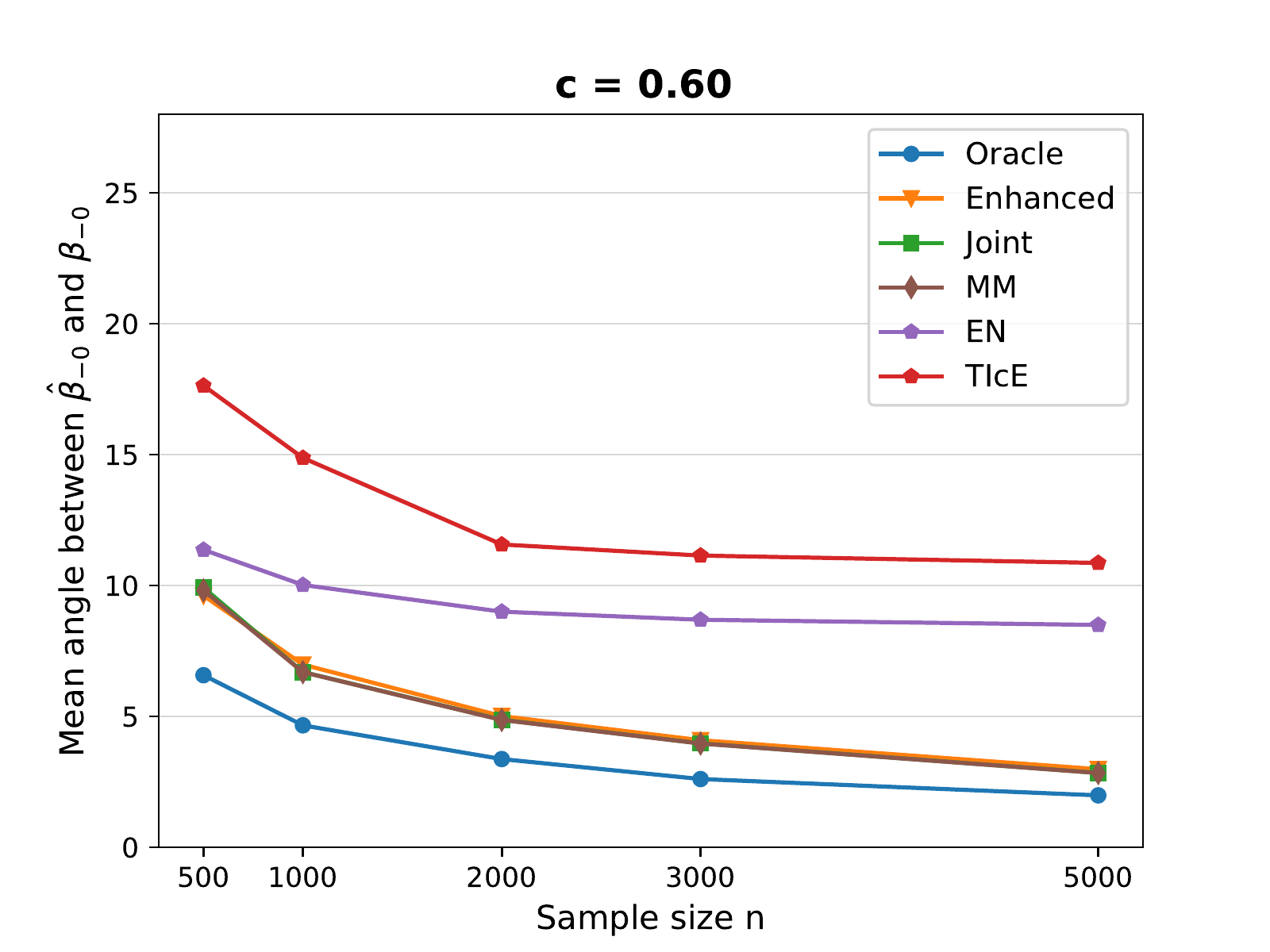}
\end{subfigure}

\hspace*{-3.8cm}
\begin{subfigure}{.6\textwidth}
    \centering   \includegraphics[scale=0.5]{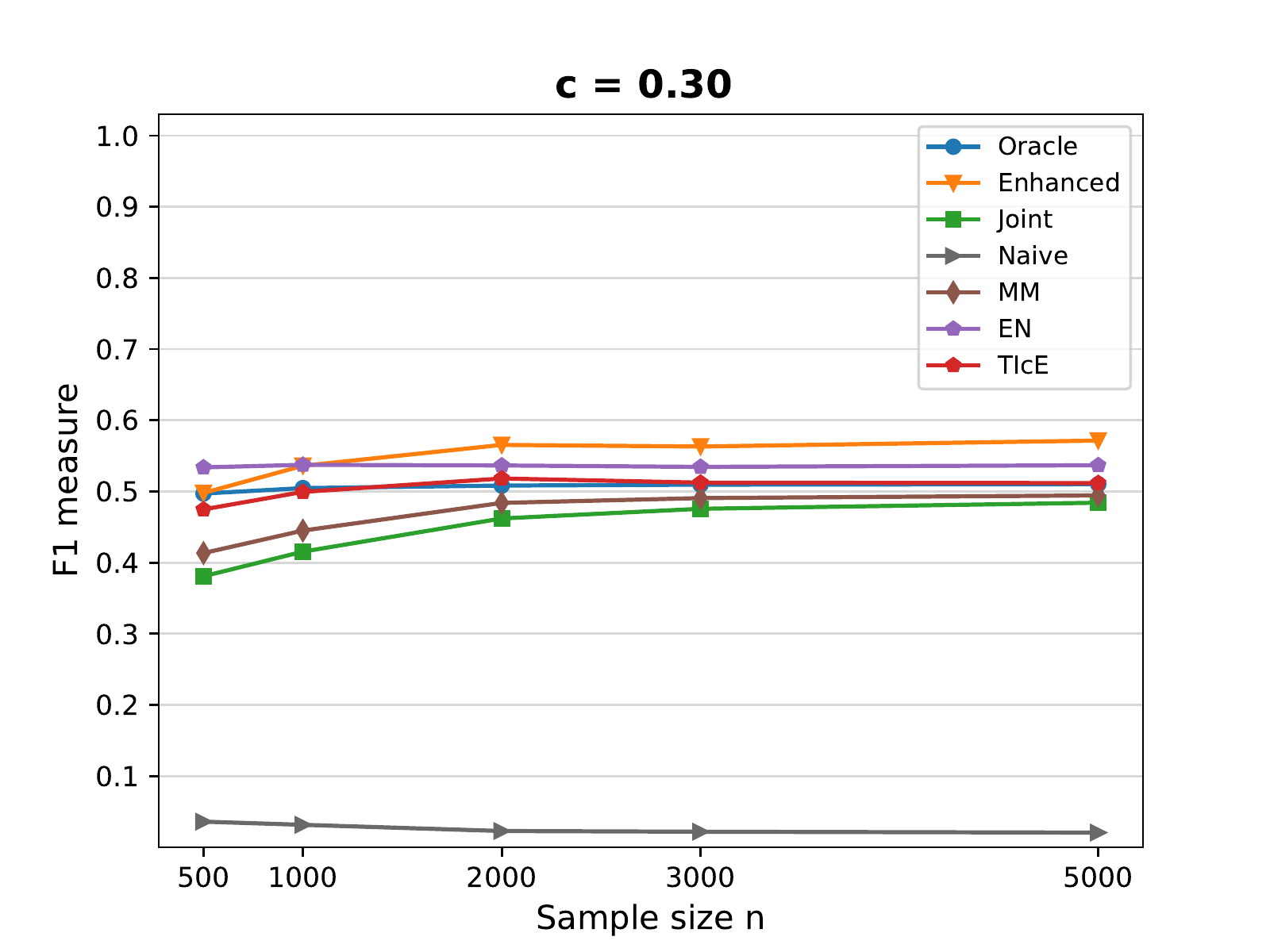}
\end{subfigure}
\hspace*{-2.3cm}
\begin{subfigure}{.01\textwidth}
    \centering    \includegraphics[scale=0.5]{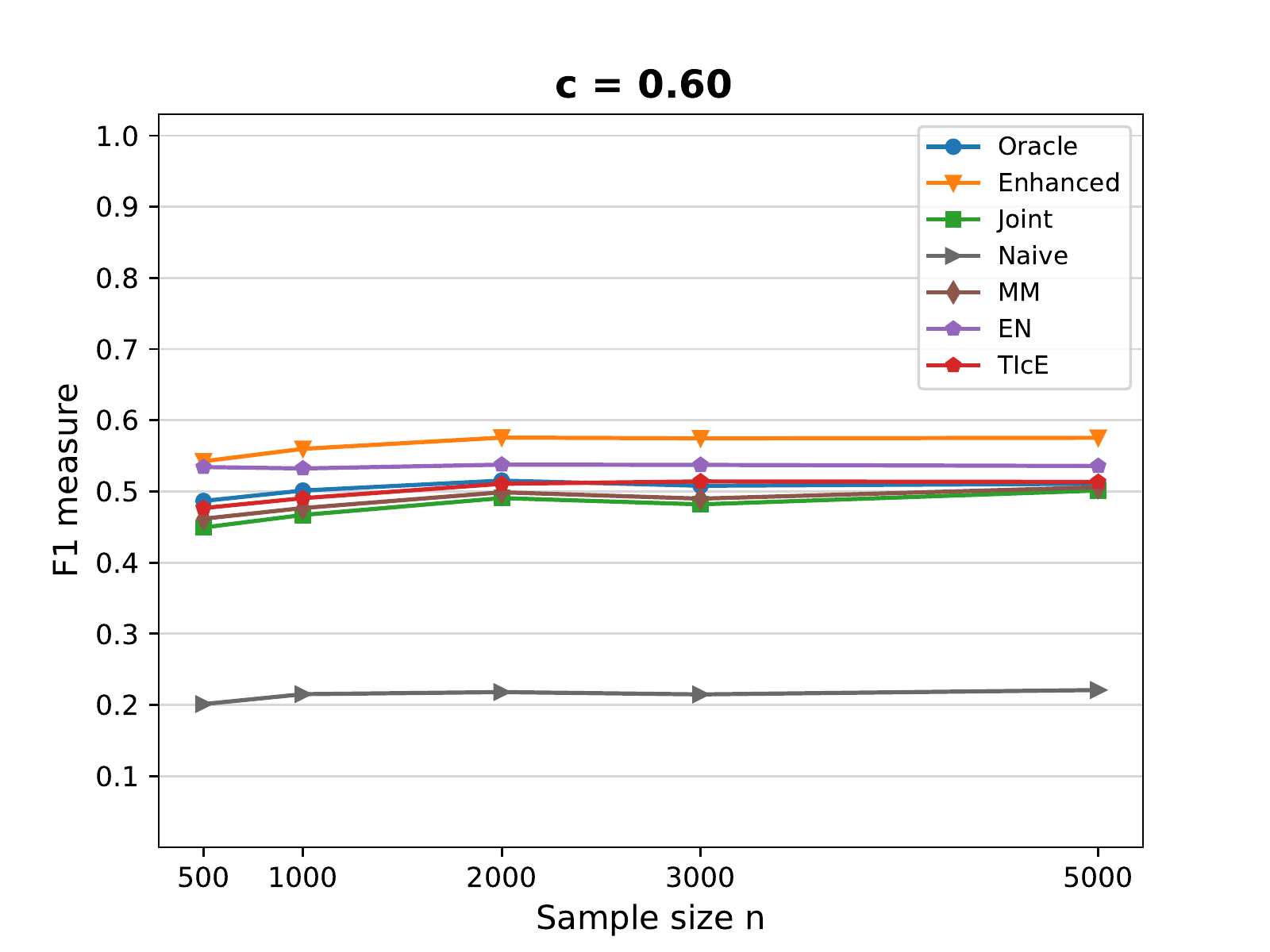}
\end{subfigure}
\hspace*{5cm}

\hspace*{-3.8cm}
\begin{subfigure}{.6\textwidth}
    \centering   \includegraphics[scale=0.5]{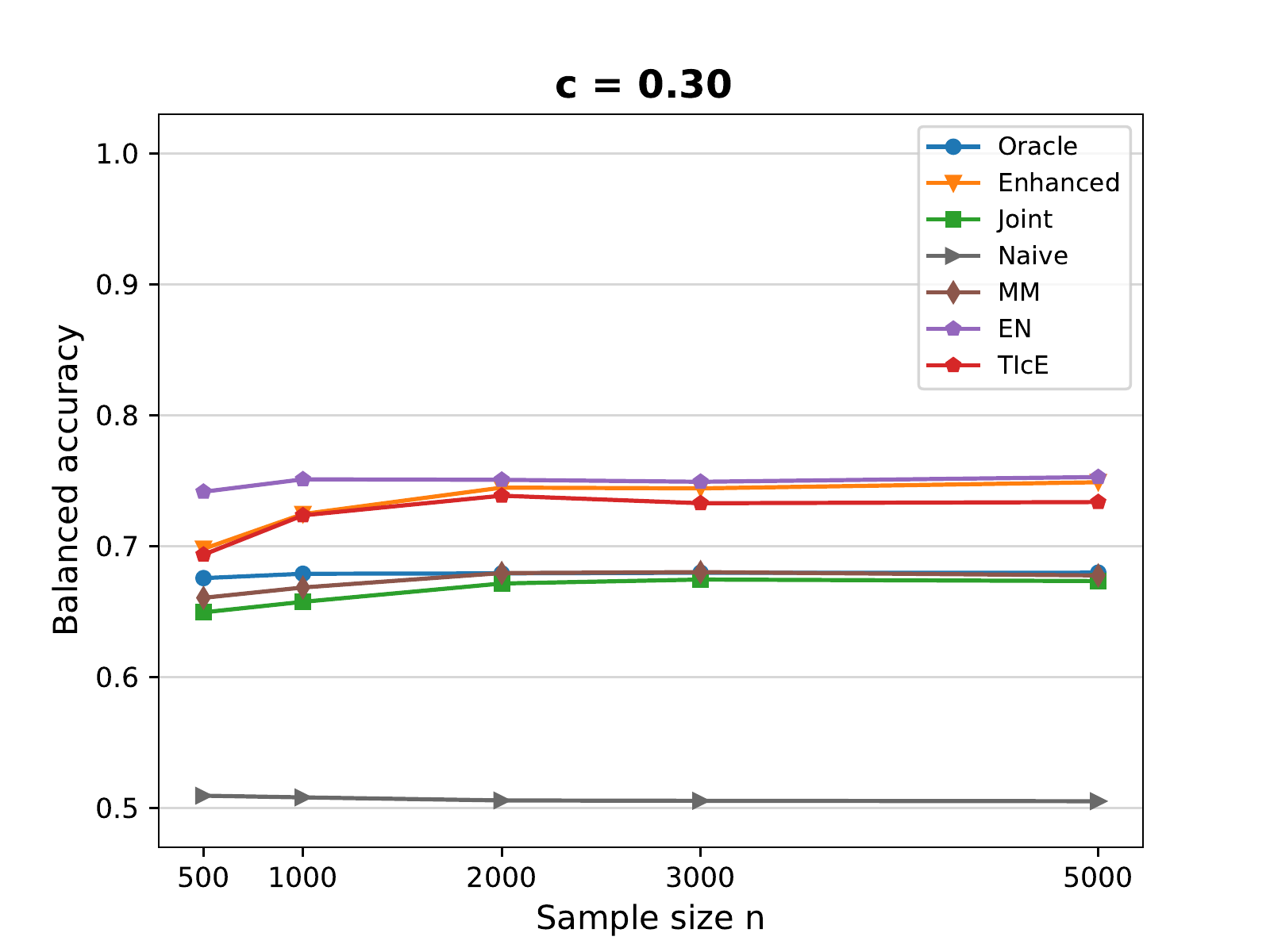} 
\end{subfigure}
\hspace*{-2.3cm}
\begin{subfigure}{.01\textwidth}
    \centering    \includegraphics[scale=0.5]{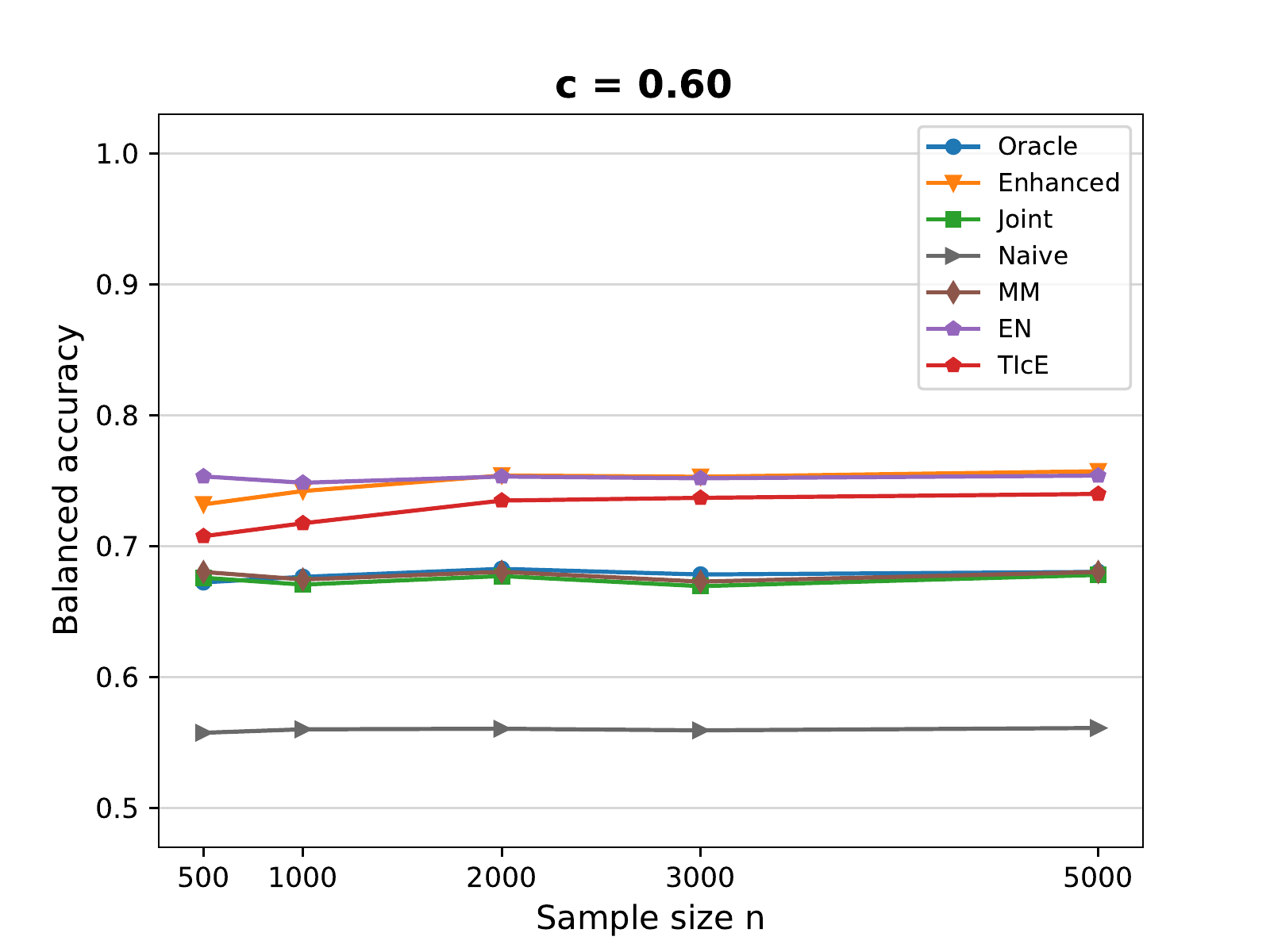}
\end{subfigure}
\hspace*{5cm}

\caption{Mean difference of angles, F1 and Balanced Accuracy against sample size for  artificial data.}
\label{artif_plot}

\end{figure*}

In order to check how Ruud's  theorem works in practice and the performance of the proposed classifier, we  considered a simple synthetic example where vector of predictors $\tilde X$  has three-dimensional normal distribution  with mean $m=(1,1,-1)^T$, variances equal to 1 and covariances $Cov(X_1, X_2) = 0.2$, $Cov(X_1, X_3) = -0.2$ and $Cov(X_2, X_3) = 0$. Thus $X_1$ is positively correlated with $X_2$ and negatively correlated with $X_3$. Moreover  posterior probability of $Y=1$ given $X=x$ is logistic with $\beta=(-1,-1,1,1)^T$.
We investigated the angle between $\hat\beta_{-0}$ and $\beta_{-0}$ for all considered estimators, the performance of corresponding classifiers for $c=0.3, 0.6$ and several values of $n$ ranging from 500 to 5000.
The results are shown in Figure \ref{artif_plot}. The first row of the panel 
exhibits goodness of fit of the considered estimators measured by the mean differences of  their angles and the angle of  $\beta_{-0}$.
It indicates that in concordance with Ruud's theorem the direction of $\beta_{-0}$ is approximately recovered by direction of
naive estimator $\hat\beta_{-0}$ for sample sizes larger than 1000 and the accuracy increases with increasing sample size. Moreover the accuracy of $\hat\beta_{-0}$ measured by mean difference of angles for naive, MM and JOINT estimators approximately coincides and is consistently better than that of EN and TIcE estimators. In terms of F1 measure shown in the second row the introduced enhanced naive classifier works consistently better  than its competitors and in terms of Balanced Accuracy (third row)  it is only outperformed by EN classifier for $c=0.3$.
\subsection{Real datasets}
\begin{figure*}[b]
\centering
\hspace*{-8.9cm}
\begin{subfigure}{.6\textwidth}
    \centering  \includegraphics[scale=0.5]{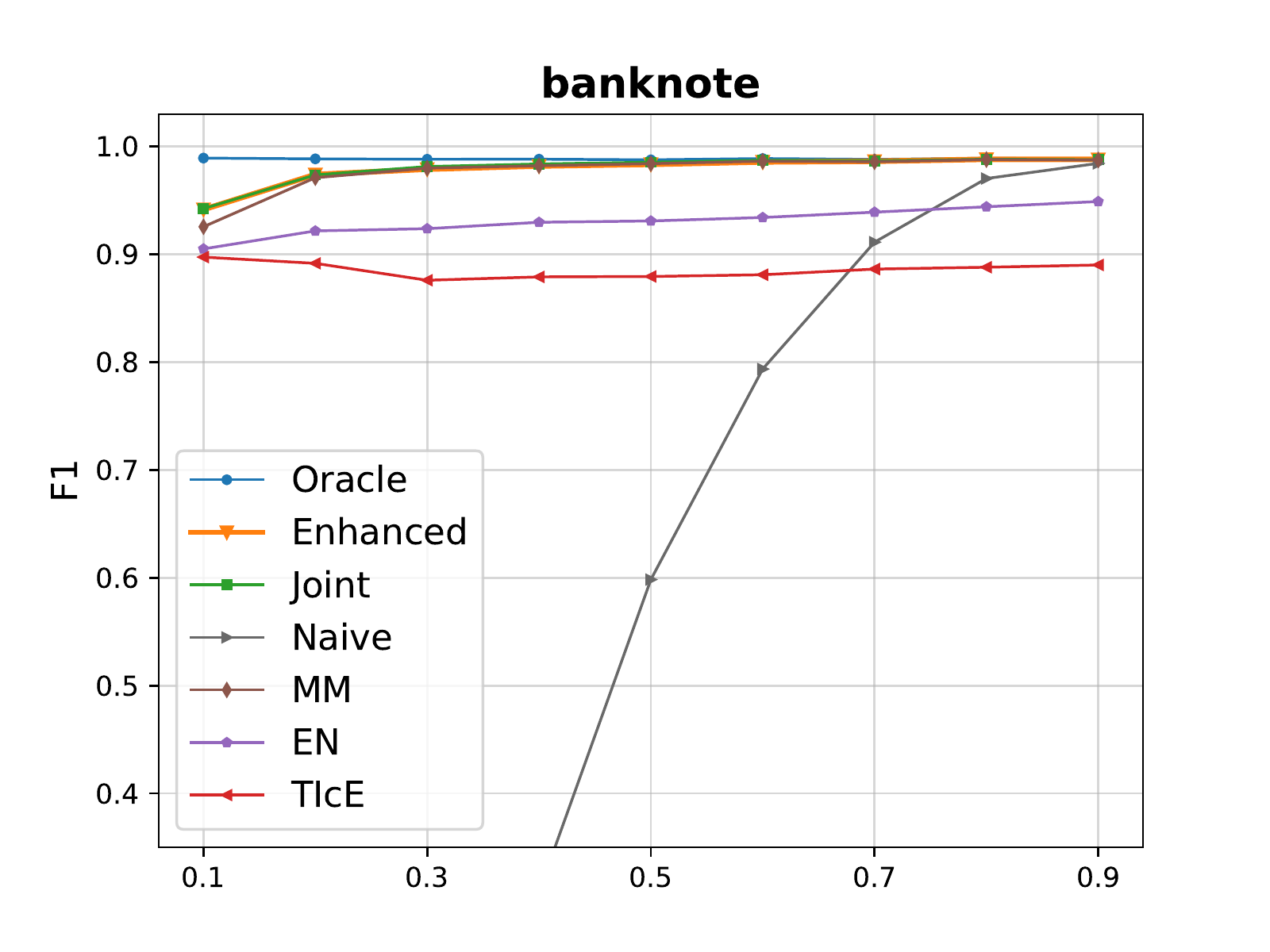}
\end{subfigure}
\hspace*{-2.3cm}
\begin{subfigure}{.01\textwidth}
    \centering \includegraphics[scale=0.5]{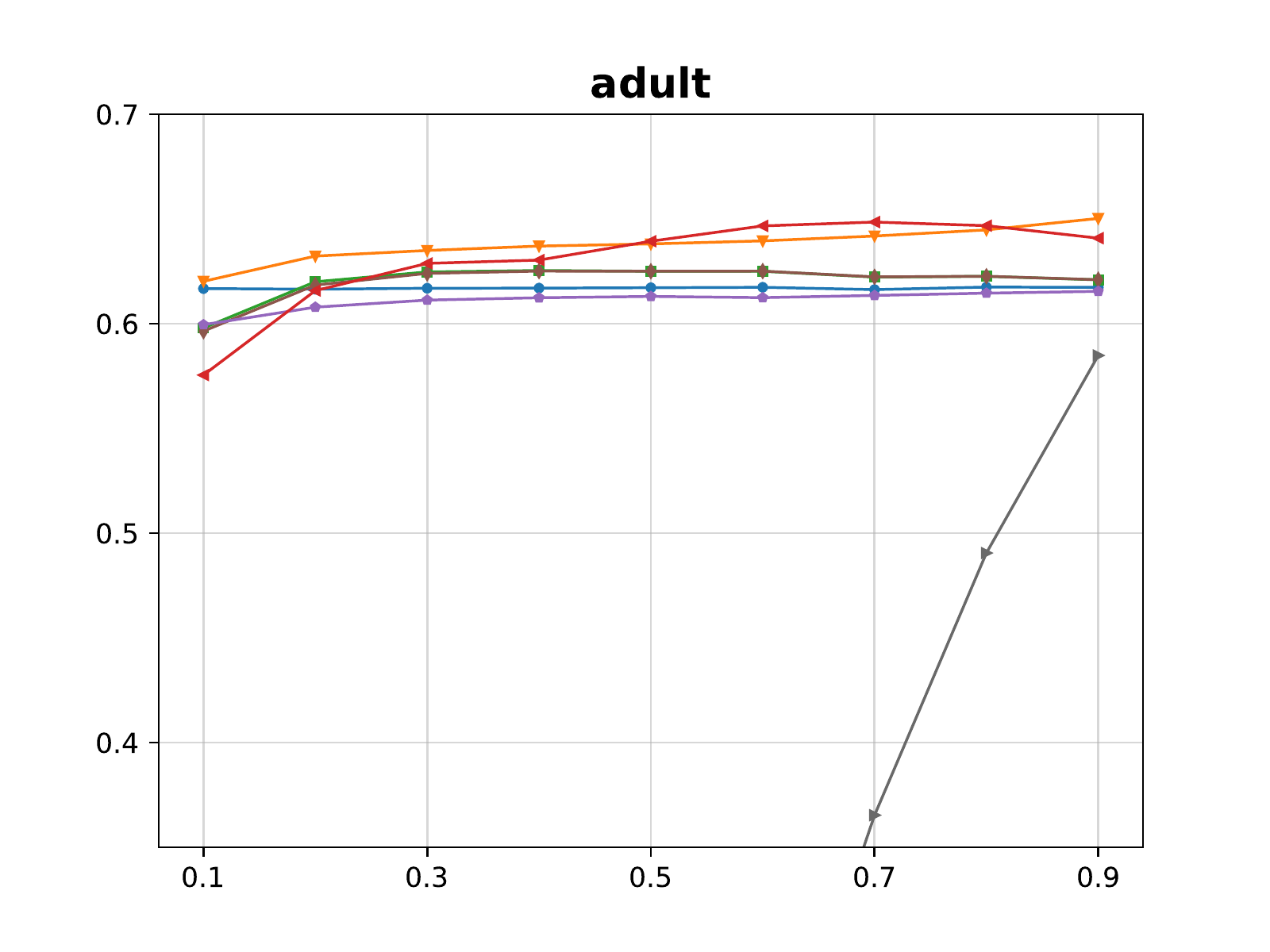}
\end{subfigure}

\hspace*{-3.8cm}
\begin{subfigure}{.6\textwidth}
    \centering   \includegraphics[scale=0.5]{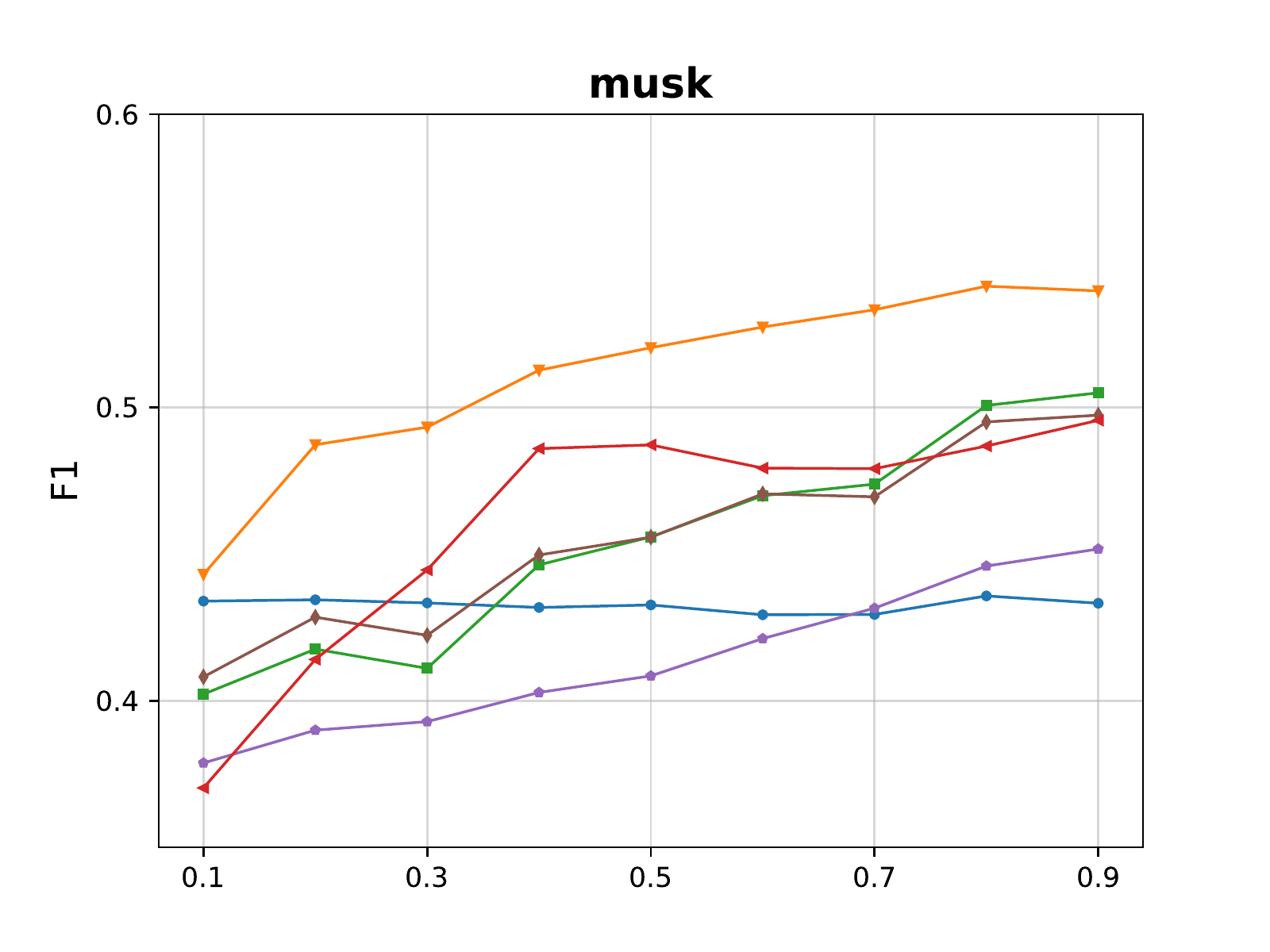}
\end{subfigure}
\hspace*{-2.3cm}
\begin{subfigure}{.01\textwidth}
    \centering    \includegraphics[scale=0.5]{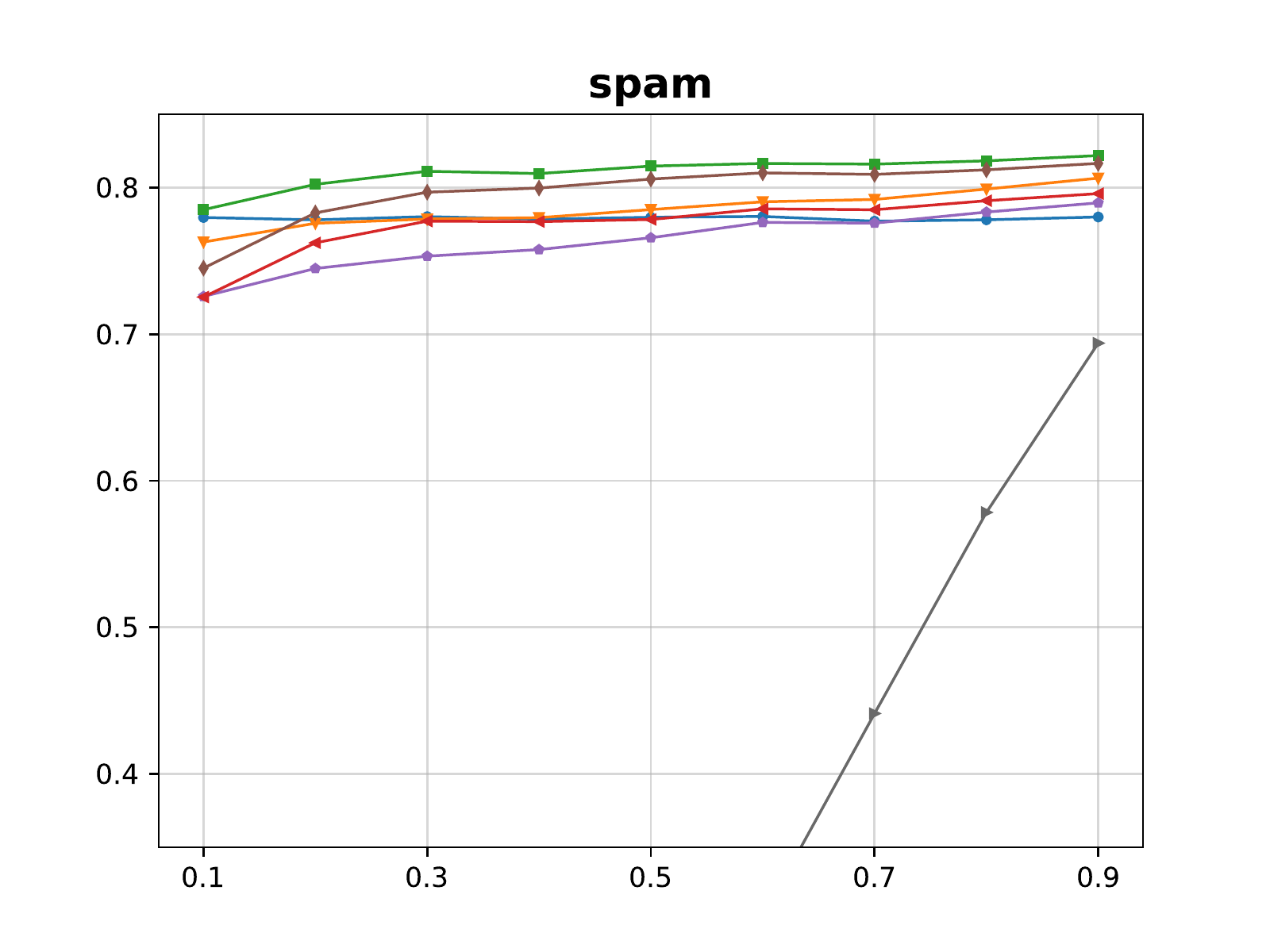}
\end{subfigure}
\hspace*{5cm}

\hspace*{-3.8cm}
\begin{subfigure}{.6\textwidth}
    \centering   \includegraphics[scale=0.5]{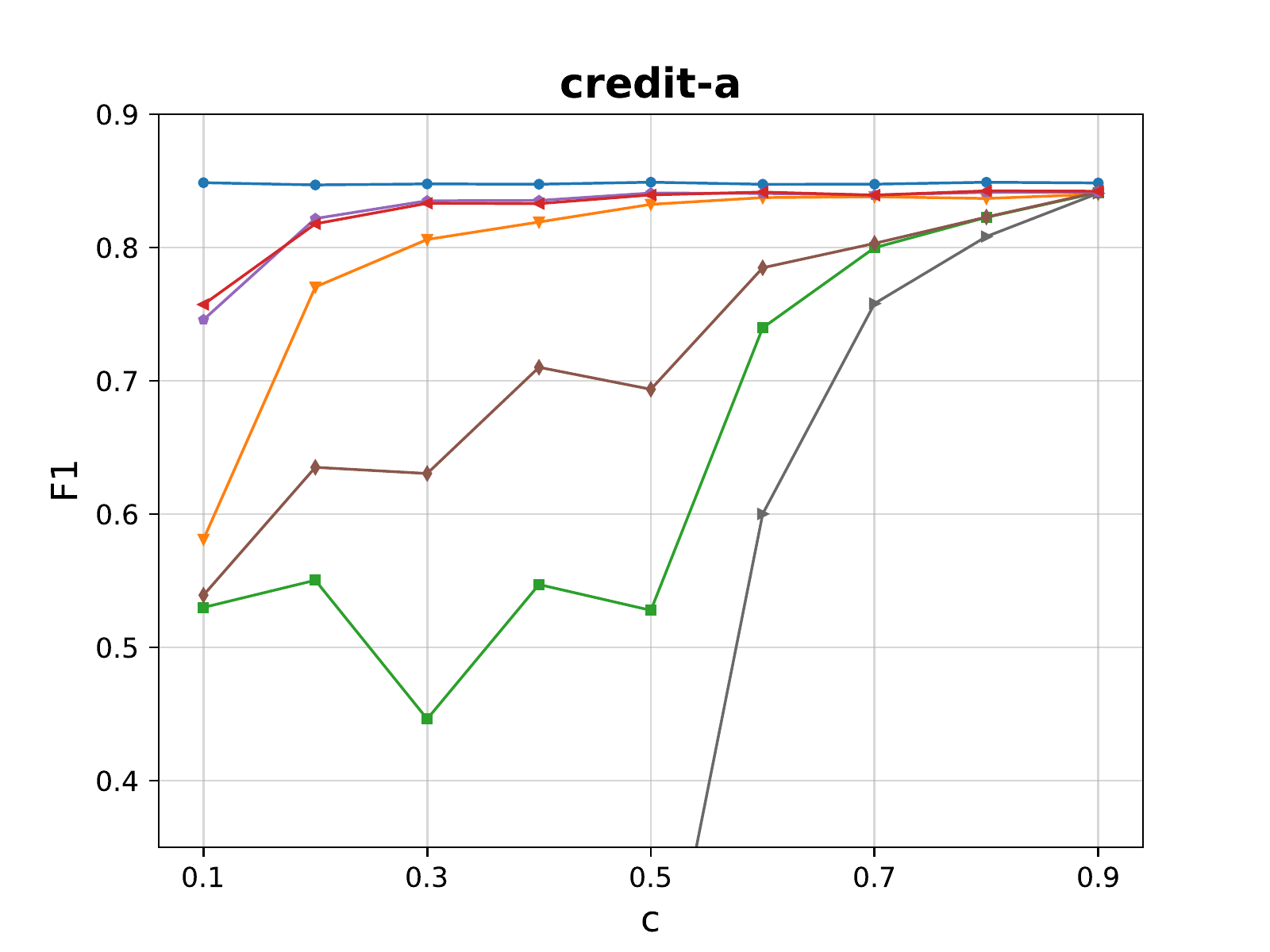} 
\end{subfigure}
\hspace*{-2.3cm}
\begin{subfigure}{.01\textwidth}
    \centering    \includegraphics[scale=0.5]{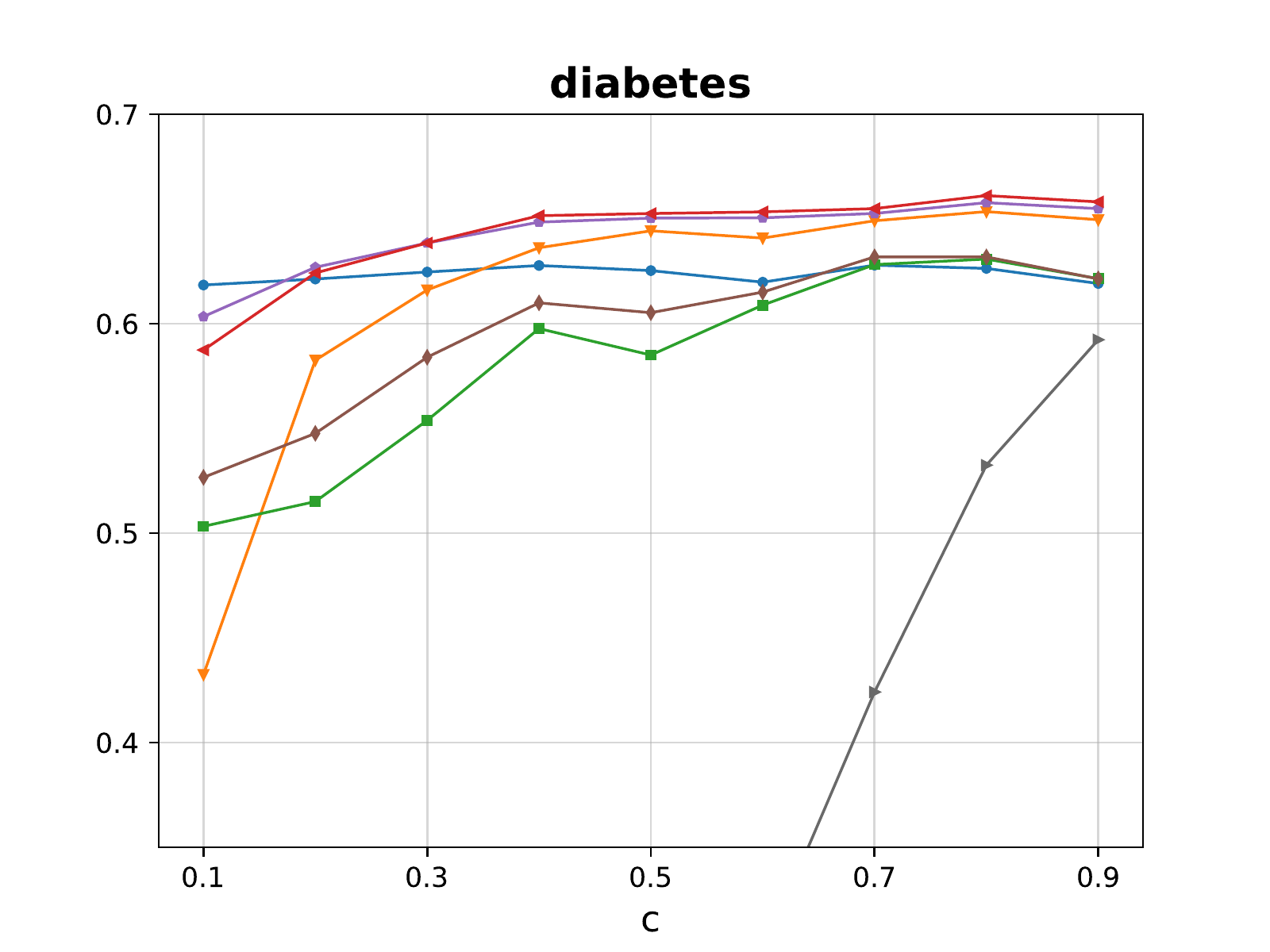}
\end{subfigure}
\hspace*{5cm}

\caption{F1 measure against values of c for the considered data sets.}
\label{F1 measure}

\end{figure*}

\begin{figure*}[b]
\centering
\hspace*{-8.9cm}
\begin{subfigure}{.6\textwidth}
    \centering  \includegraphics[scale=0.5]{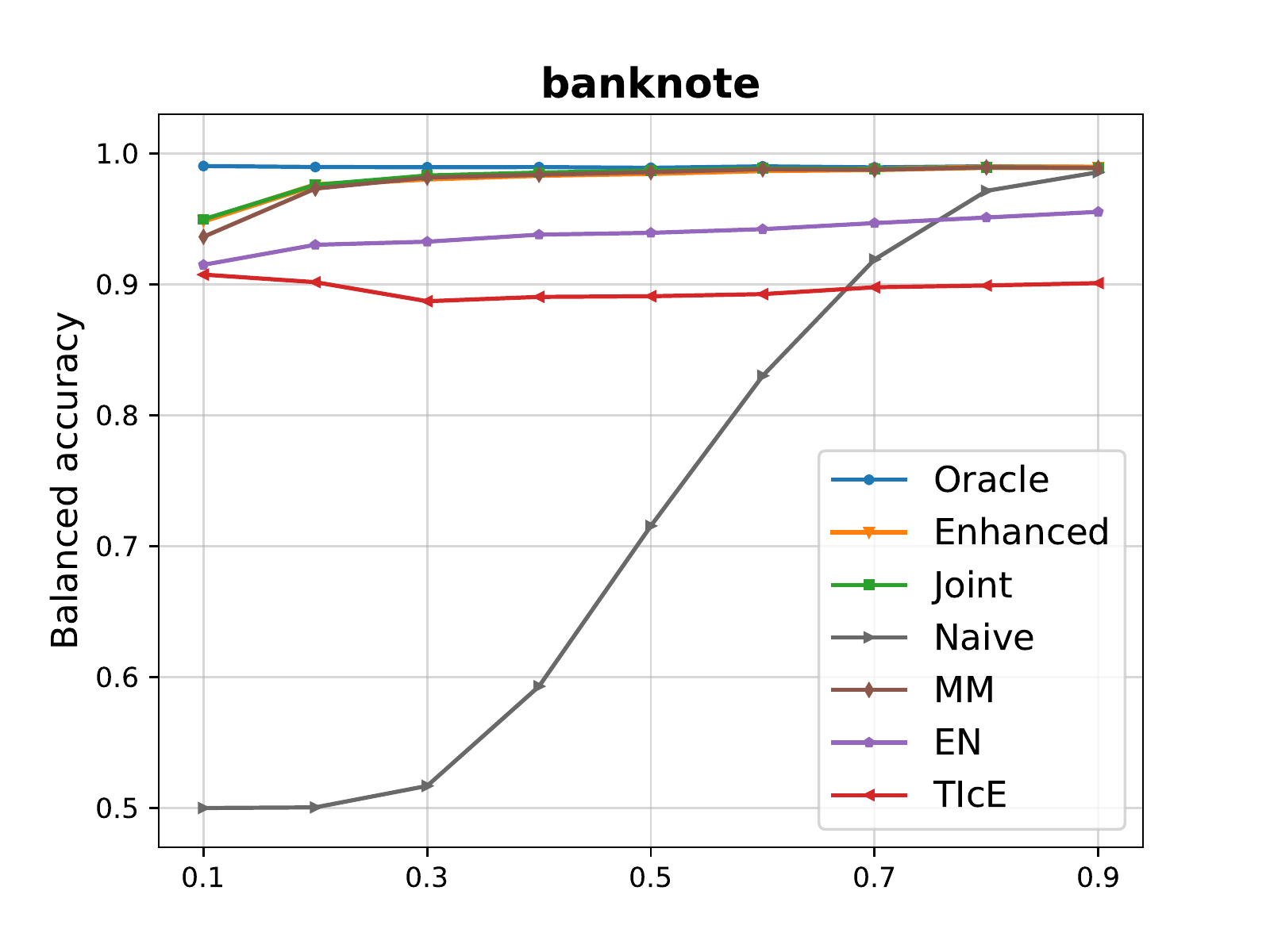}
\end{subfigure}
\hspace*{-2.3cm}
\begin{subfigure}{.01\textwidth}
    \centering \includegraphics[scale=0.5]{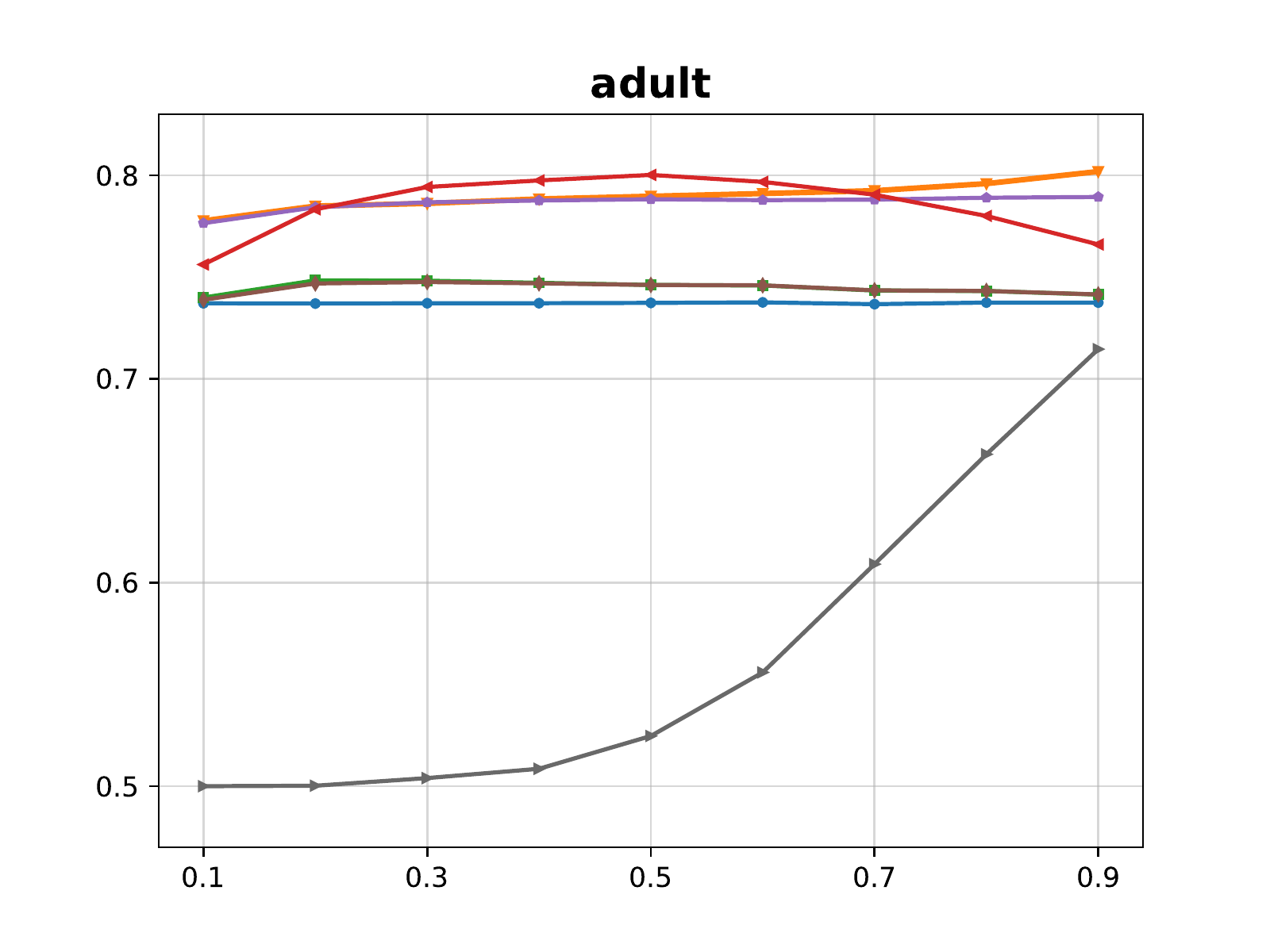}
\end{subfigure}

\hspace*{-3.8cm}
\begin{subfigure}{.6\textwidth}
    \centering   \includegraphics[scale=0.5]{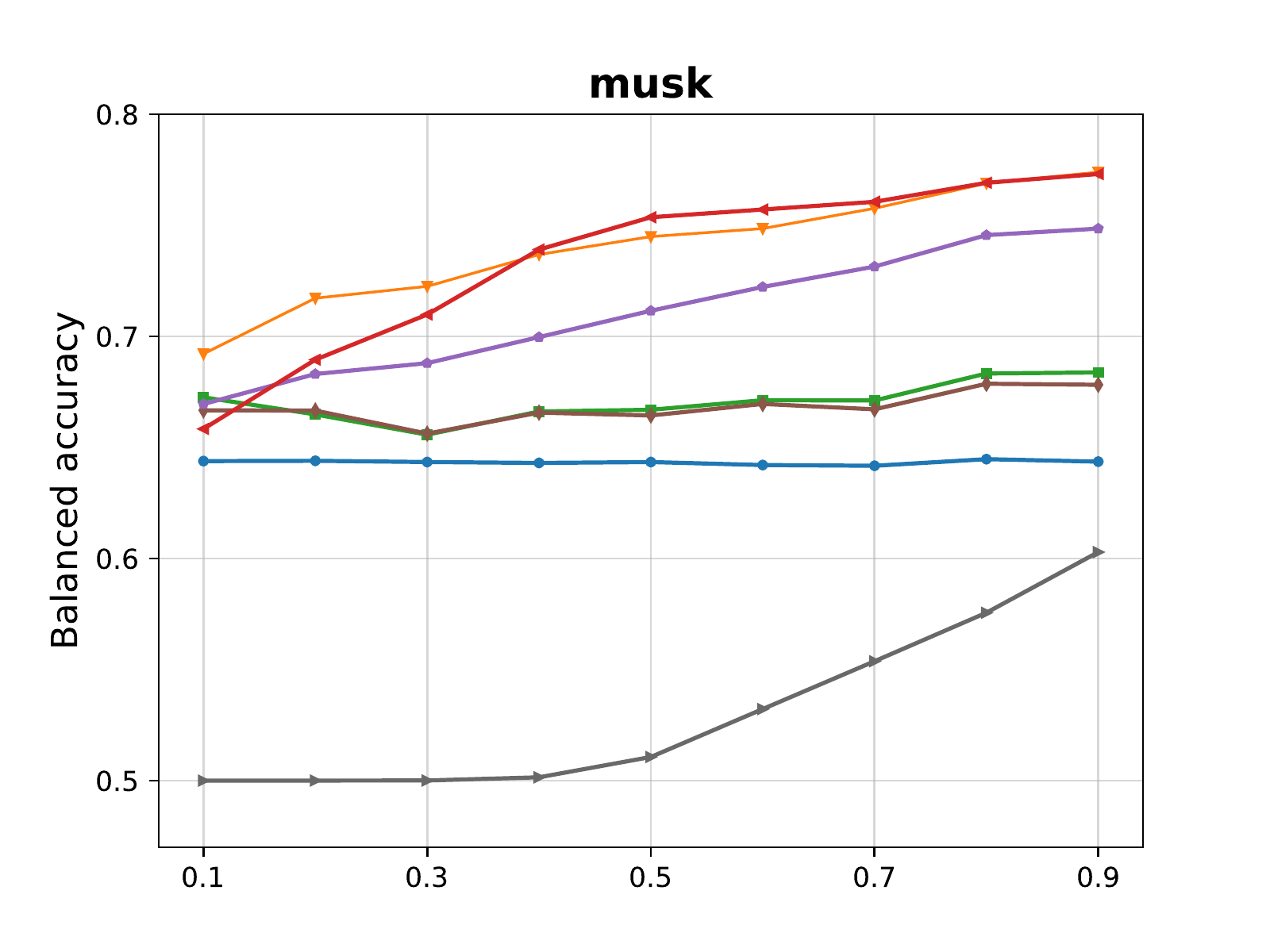}
\end{subfigure}
\hspace*{-2.3cm}
\begin{subfigure}{.01\textwidth}
    \centering    \includegraphics[scale=0.5]{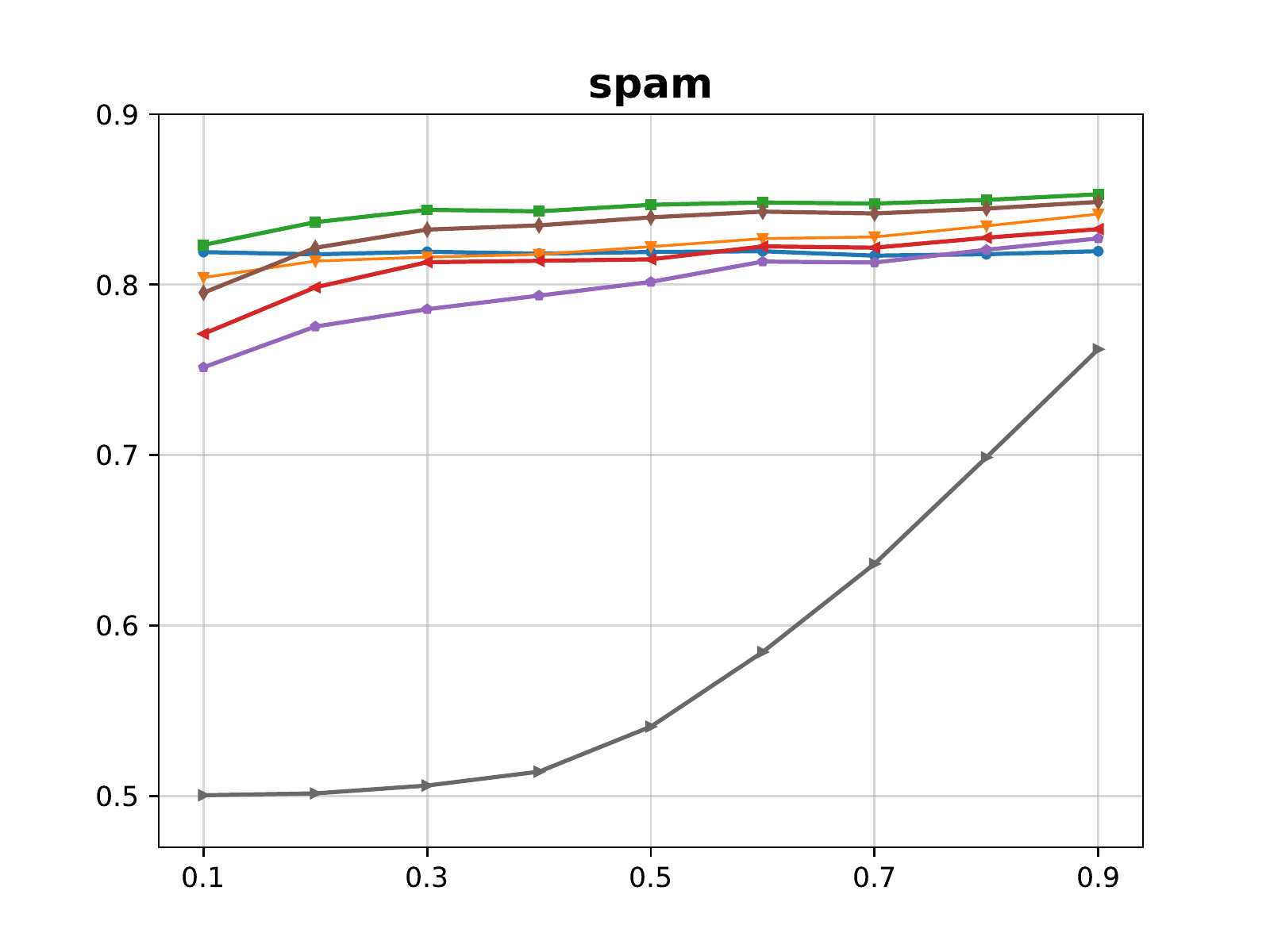}
\end{subfigure}
\hspace*{5cm}

\hspace*{-3.8cm}
\begin{subfigure}{.6\textwidth}
    \centering   \includegraphics[scale=0.5]{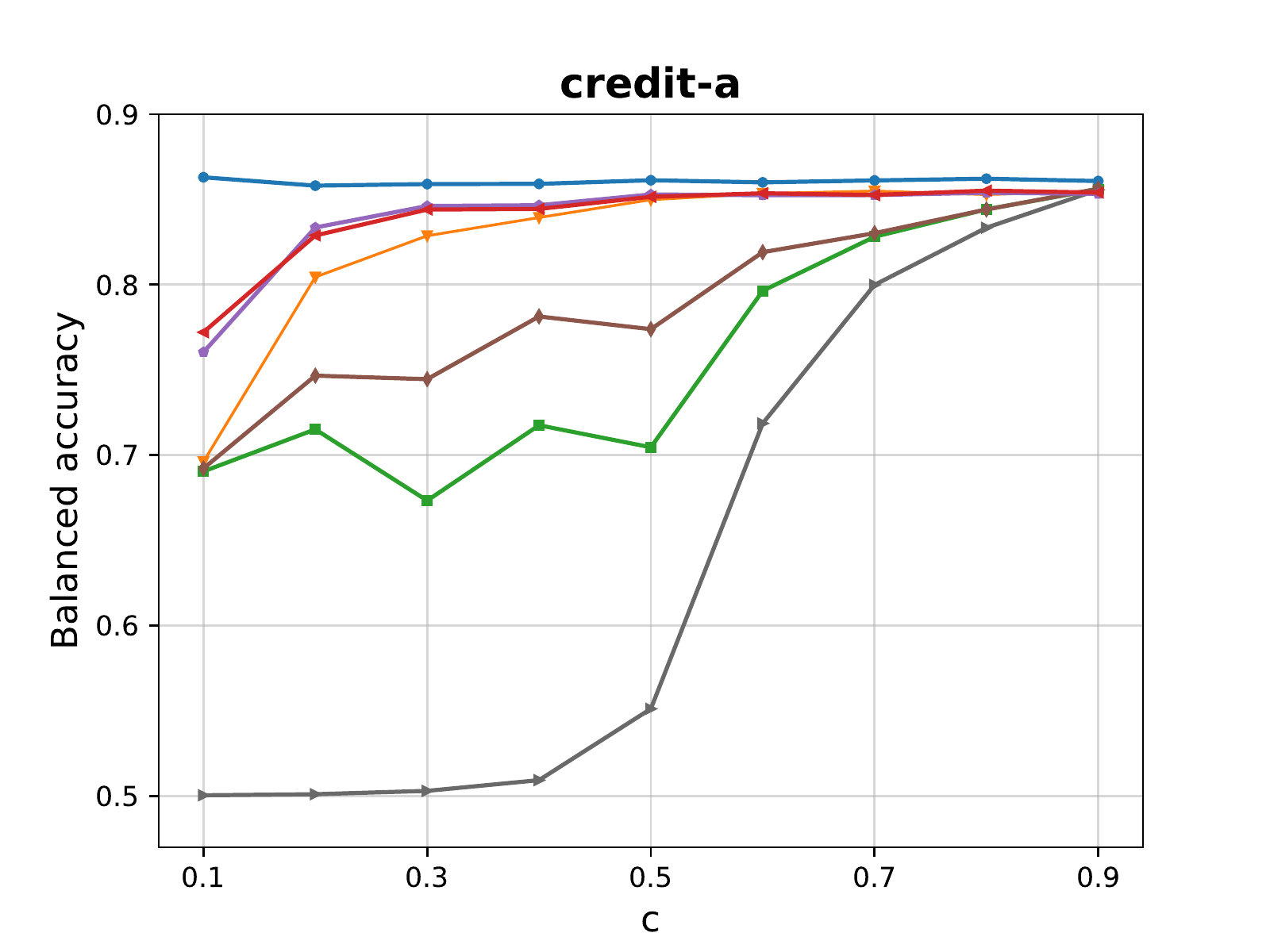} 
\end{subfigure}
\hspace*{-2.3cm}
\begin{subfigure}{.01\textwidth}
    \centering    \includegraphics[scale=0.5]{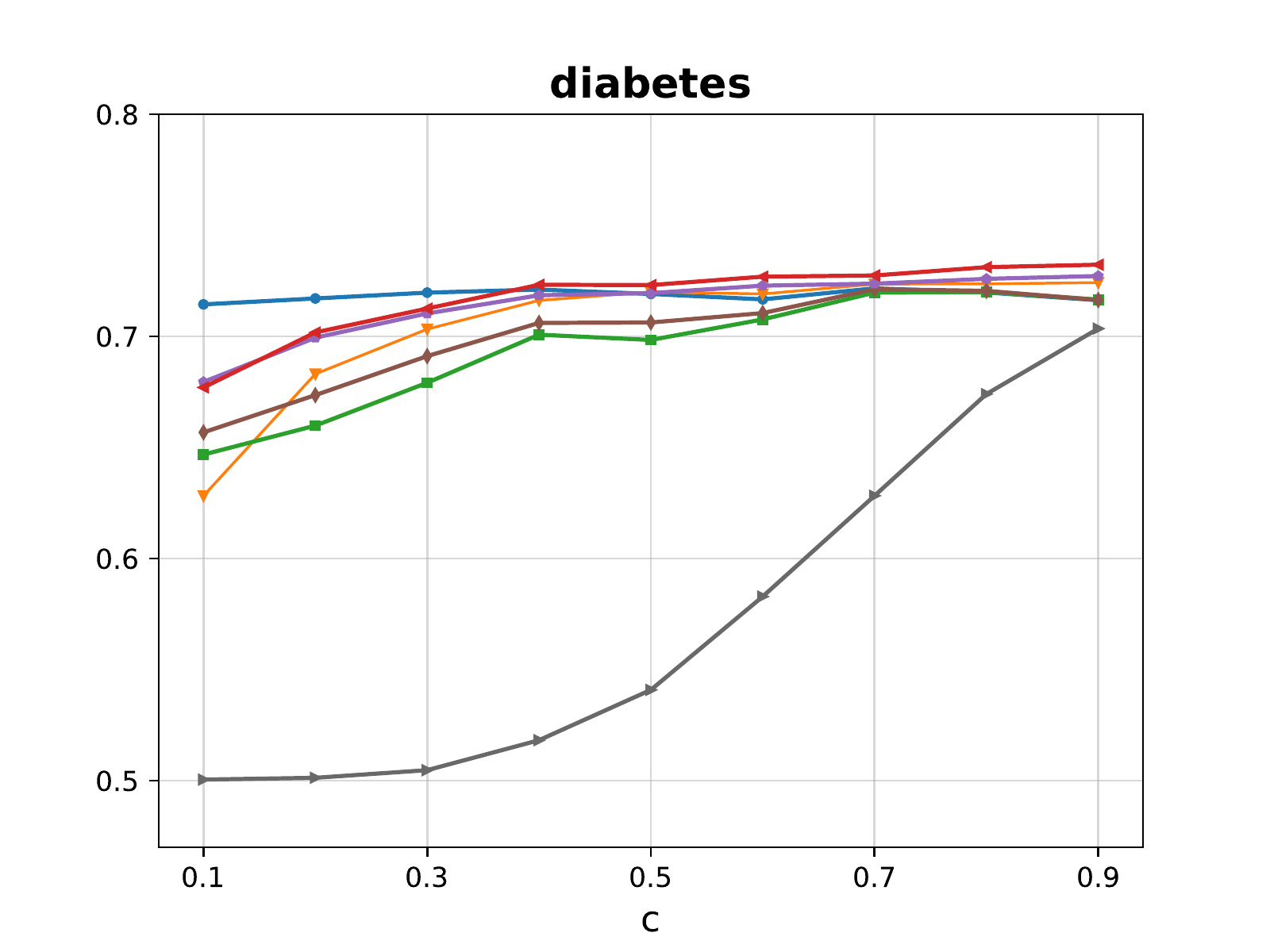}
\end{subfigure}
\hspace*{5cm}

\caption{Balanced Accuracy against values of c for the considered data sets.}
\label{Bal_acc}

\end{figure*}

We have analysed performance of the estimators on six data sets from UCI directory with sample sizes ranging from around 300 to 30 000 and number of features from 3 to 166 (the main characteristics of the data sets are given in Table \ref{zbiory_danych}). The figures show mean performance with the regard of F1 measure (Figure \ref{F1 measure}) and Balanced Accuracy (Figure \ref{Bal_acc}), for values of $c$ ranging from $0.1$ to $0.9$, based on 200 random splits of the data into training and testing subsamples. Standard errors for the mean are smaller than 0.01 in most cases for both F1 and BA measure with the only exception of F1 measure on {\tt credit-a} and {\tt diabetes} data set and the maximal value of SE is 0.026 for JOINT estimator on {\tt credit-a}.
Note that the results for the naive classifier are truncated from below in Figure \ref{F1 measure}: F1 measure for naive classifier is very low for $c\leq 0.5$ and approach 0 for $c$ close to 0. The  first immediate observation is that the change of the intercept estimator, which is the only difference between the  naive classifier and its enhanced version, has a huge impact on its performance with regard to both considered measures.\\\\
{\bf F1 measure} In all cases but one case the enhanced classifier works better (data sets {\tt musk, credit-a, diabetes, adult}) or on par ({\tt banknote}) with JOINT and MM estimators. In the case of {\it spam} it works marginally worse than JOINT and MM. 
This is interesting, especially  in comparison with  MM estimator which requires much more computing effort. It also outperforms TIcE and EN estimators on three data  sets: {\tt banknote, musk} and {\tt spam}. On {\tt adult} data set enhanced classifier works better than EN and on par with TIcE.
Its excellent performance on {\tt musk} data set is worth pointing out.
The performance of enhanced estimator deteriorates for  small values of $c$, possibly due to  loss of accuracy of $\widehat{F1}_{PU}$ (note that the  denominator of (\ref{F1PU2})  becomes smaller for smaller $c$).\\\\
{\bf Balanced Accuracy} The performance of enhanced estimator with respect of Balanced Accuracy is similar to that with respect to F1 measure.\\
We have also analysed training times of the considered classifiers. Table \ref{czas} shows the training times for the largest data set {\tt adult}. In the case of Enhanced and JOINT classifiers the times are approximately the same and 2-3 times shorter that the times for EN and TiCE classifiers.  The most computation intensive is  MM classifier as it requires inner loop of convex optimisation for each iteration of $\hat\beta$.

\section{Conclusion}
We have studied a novel  modification of naive classifier for Positive Unlabeled data under SCAR assumption. The classifier has strong theoretical underpinnings  following from  Ruud's theorem which are are established in Theorem \ref{Ruud}. These indicate that the coefficients of logistic classifier corresponding to genuine predictors are consistently estimated based on observed $(X,S)$ data  and the estimation problem boils down to
consistent estimator of the intercept. We have proposed such an estimator based on maximisation of observable analogue of $F1$ measure. Moreover, we have shown analysing real data sets that the resulting enhanced naive  estimator is a promising alternative to classifiers based on parametric models of posterior probability (JOINT and MM classifier) as well as nonparametric ones (TIcE and EN classifiers). Future research may include finding alternatives to the proposed method of estimating the intercept as well as extension of the considered method to the situation when SCAR assumption is violated. In particular, note that when posterior probability $y(x)$ satisfies (\ref{logistic}) and $e(x)$ is an {\it arbitrary} function of $y(x)$, posterior probability  $s(x)$ of $S=1$ given $X=x$ is a function of $x^T\beta$ and it corresponds to misspecified logistic model. Thus the conclusion of Theorem \ref{Ruud} applies  also  to this more general situation which as its special case includes probabilistic gap assumption when $e(x)$ is an increasing function of $y(x)$.

\begin{table}[]
    \centering
    \begin{tabular}{l|c|c|c}
    Name & Size & Features & Fraction of positive observations \\\hline
    adult & 32561 & 57 & 0.24 \\
    banknote & 1372 & 4 & 0.44 \\
    breast-w & 699 & 9 & 0.34 \\
    credit-a & 690 & 38 & 0.44 \\
    diabetes & 768 & 8 & 0.35 \\
    haberman & 306 & 3 & 0.26 \\
    ionosphere & 351 & 34 & 0.36 \\
    musk & 6598 & 166 & 0.15 \\
    spambase & 4601 & 57 & 0.39 \\
    
    \end{tabular}
    
    \caption{Analysed datasets and their statistics}
    \label{zbiory_danych}
    \end{table}

\begin{table}[]
    \centering
    \begin{tabular}{c|c|c|c|c|c|c}
    Algorithm & Oracle & Enhanced & JOINT & MM & EN & TIcE \\\hline
    Time & 0.05s & 0.22s & 0.23s & 201s & 0.66s & 0.9s \\
    \end{tabular}

    \caption{Mean training time in seconds on the largest dataset adult with $c = 0.5$.}
    \label{czas}
    \end{table}

\vskip 1cm



\balance




\bibliography{References_ENH}
\bibliographystyle{plain}

\end{document}